\documentclass{article}

 \usepackage[preprint]{neurips_2026}

\usepackage[utf8]{inputenc} 
\usepackage[T1]{fontenc}    
\usepackage{hyperref}       
\usepackage{url}            
\usepackage{booktabs}       
\usepackage{amsfonts}       
\usepackage{nicefrac}       
\usepackage{microtype}      
\usepackage{xcolor}         
\usepackage{amsmath}
\usepackage{graphicx}
\usepackage{wrapfig}
\usepackage{multirow}

\title{SurgWMBench: A Vision-Based Benchmark for World-Modeling Surgical Instrument Motion Planning}

\author{
\textbf{Huanrong Liu}$^{1}$\thanks{Equal contribution.}
\ \ \ 
\textbf{Weiliang Huang}$^{1}$\footnotemark[1]
\ \ \ 
\textbf{Bob Zhang}$^{1}$\thanks{Corresponding authors.}
\ \ \ 
\textbf{Weichao Cai}$^{3}$
\\
\textbf{Chunlin Tian}$^{2}$
\ \ \ 
\textbf{Qingbiao Li}$^{1}$\footnotemark[2]
\\[1ex]
$^{1}$University of Macau
\ \ \ 
$^{2}$National University of Singapore
\ \ \ 
$^{3}$Xiamen University
\\[1ex]
\texttt{bobzhang@um.edu.mo}
\ \ \ 
\texttt{qingbiaoli@um.edu.mo}
}

\begin{document}

\maketitle



\begin{abstract}
Reliable surgical planning requires models that move beyond recognizing the current surgical step or imitating expert demonstrations, and instead anticipate how instrument motion reshapes subsequent operative states. Most surgical video understanding methods focus on recognizing phases, actions, or workflow states, while providing limited support for explicitly modeling instrument motion. Conversely, existing tool motion prediction methods can forecast instrument trajectories, but they generally do not capture the coupled evolution of future surgical video states. World models offer a natural framework for jointly modeling visual state transitions and instrument motion dynamics. However, existing surgical world model studies remain largely centered on visual generation quality, relying on generation-oriented metrics such as FVD and CD-FVD. These metrics are poorly aligned with instrument motion planning, as they do not directly measure whether predicted trajectories are geometrically accurate, temporally coherent, or actionable for downstream planning. This limitation is partly structural, since the field lacks public datasets and standardized evaluation protocols that provide the benchmarking infrastructure needed to assess motion-centric capabilities in surgical world models. In this paper, we introduce SurgWMBench, a vision-based benchmark for short-horizon surgical motion planning and dynamics prediction. Given intraoperative image sequences and historical instrument trajectory, SurgWMBench evaluates both near-future instrument motion prediction and stability under continuous rollout or input perturbations. We further design a hierarchical evaluation protocol to assess the capability boundaries of existing paradigms from multiple perspectives, including fine-grained dynamics modeling, continuous rollout stability, and perturbation recovery. This benchmark provides a unified platform for reproducible evaluation of surgical world models and lays the foundation for building safer and more reliable vision-based surgical motion planning systems.
\end{abstract}
\section{Introduction}

Surgical planning is a core decision-making problem in intelligent surgical systems, spanning multiple levels from high-level workflow understanding to low-level tool motion generation, trajectory optimization, and risk control \cite{doi:10.1126/scirobotics.abj2908,attanasio2021autonomy}. It is a key capability for advancing surgical robots from passive execution tools toward proactive intelligent systems. For surgical robots in clinical scenarios, reliable planning requires models not only to understand the current intraoperative state, but also to predict how instrument motion will change the subsequent operative field and to maintain stability and safety during continuous execution.

With the development of surgical video learning, robotic imitation learning, and sequence modeling methods, substantial progress has recently been made in phase recognition, action anticipation, workflow prediction, and trajectory learning \cite{twinanda2016endonet, nwoye2022data, wang2022autolaparo, wagner2023comparative, yuan2021surgical, shi2022recognition}. 
However, these efforts have largely progressed along separate tracks, leaving instrument motion planning insufficiently explored. Most surgical video understanding methods~\cite{twinanda2016endonet, wagner2023comparative} focus on recognizing phases, actions, or workflow states, but do not explicitly reason about how instruments should move within the operative field. In contrast, existing motion planning methods~\cite{shi2022recognition,hansen2026imitatecholec} can forecast future instrument trajectory, yet they often treat motion as an isolated signal and do not model the corresponding evolution of future surgical video states. Simply predicting instrument motion is insufficient for clinical surgery, since a planning model must also understand the visual consequences of motion, including how instruments interact with tissues, how tissues respond, and how the local operative field changes over time. Therefore, surgical planning should move beyond recognizing the current step or reproducing expert trajectories, and should instead require the joint modeling of future surgical video states and instrument motion trajectories.

\begin{figure}[t]
    \centering
    \includegraphics[width=\linewidth]{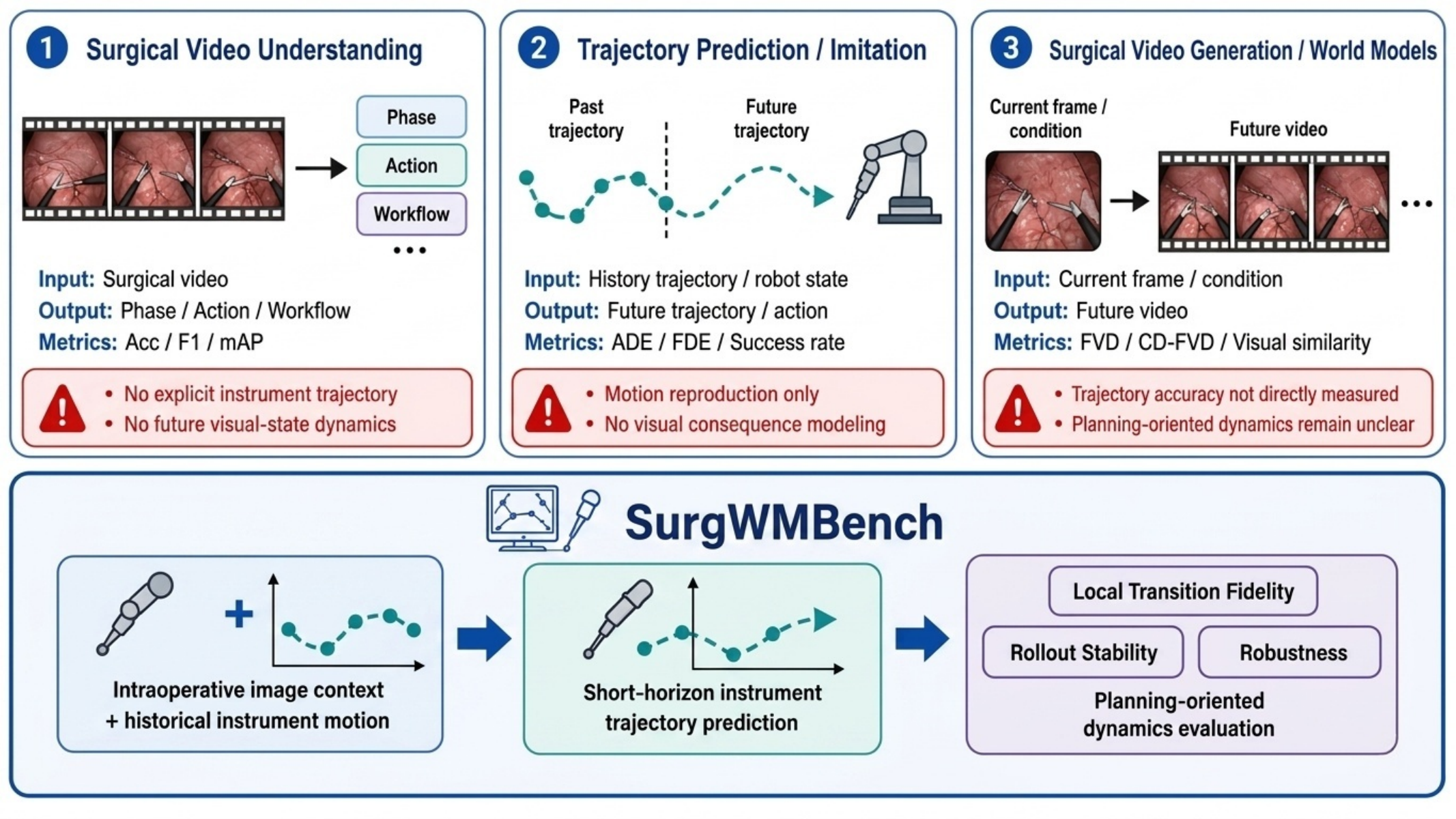}
    \caption{
Existing surgical benchmarks mainly evaluate semantic video understanding, trajectory reproduction, or video generation quality, while reliable surgical planning requires planning-oriented short-horizon visual-motion dynamics evaluation. SurgWMBench fills this gap by assessing trajectory-level accuracy, closed-loop rollout stability, and robustness under perturbations.
    }
    \label{fig:motivation}
    \vspace{-4.5mm}
\end{figure}

In this sense, trustworthy surgical intelligence depends on predictive modeling of how the surgical process continuously evolves under actions, which constitutes the broader problem of surgical world modeling \cite{ha2018recurrent, bruce2024genie}. Some recent studies have begun to introduce world-modeling ideas into surgical visual scenes, exploring future video generation, action-conditioned generation, or interactive simulation to model the temporal evolution of surgical procedures \cite{koju2025surgicalvision, chen2025surgsora, rapuri2026saw, he2025surgworld}. These studies suggest that world models have the potential to connect visual state prediction with surgical action modeling, providing a new technical pathway for prospective planning, candidate action comparison, and robotic policy learning.

However, existing surgical world model studies still mainly focus on the visual quality, temporal realism, or downstream utility of generated videos. This emphasis is partly driven by the lack of public datasets and standardized evaluation protocols that explicitly support motion-centric assessment of surgical world models. As a result, common metrics such as FVD \cite{unterthiner2018towards}, CD-FVD \cite{ge2024content}, or visual similarity scores can measure whether generated videos are close to real data in overall appearance and motion distribution. However, they provide limited evidence about whether the instrument trajectories in the generated results are accurate, whether local state transitions follow real surgical dynamics, or whether the model can support reliable motion planning. Therefore, although current surgical world models demonstrate the potential to generate future surgical scenes, their planning-oriented dynamics modeling capability still lacks direct, unified, and reproducible evaluation. As illustrated in \autoref{fig:motivation}, existing surgical benchmarks mainly evaluate semantic video understanding, motion reproduction, or video generation quality, leaving short-horizon planning-oriented dynamics underexplored.

To fill this gap, we propose SurgWMBench, a vision-based surgical motion planning benchmark for short-horizon surgical dynamics. Unlike traditional motion planning benchmarks that only evaluate predicted trajectory of motion, SurgWMBench is designed to serve surgical world model. It focuses on whether a model can predict short-term instrument motion within intraoperative image contexts, thereby providing a quantitative evaluation basis for future video-state modeling and closed-loop motion planning. Unlike existing surgical world model studies that primarily evaluate video generation quality, SurgWMBench extends the evaluation focus from visual generation quality to trajectory-level dynamic consistency. It systematically assesses model performance in motion planning accuracy, continuous rollout stability, and perturbation recovery, complementing existing video generation evaluations in motion-oriented dynamics modeling.

Specifically, SurgWMBench is constructed from real robot-assisted surgical videos with annotations centered on surgical motion. Each motion segment is uniformly sampled into 20 frames, and a 2D instrument anchor point is annotated in each frame, forming a 20-point instrument trajectory. Based on these annotations, we further design a hierarchical evaluation protocol to assess surgical world modeling capability from three complementary perspectives: 1) Local Transition Fidelity (LTF) measures whether local state transition prediction is accurate; 2) Closed-loop Rollout Stability (CRS) measures error accumulation and trajectory stability during continuous autoregressive prediction; and 3) Robustness under Perturbation and Shift (RPS) measures model reliability under perturbed inputs and distribution changes.

Therefore, our contributions can be summarized as follows.
\begin{itemize}
    \item We introduce SurgWMBench, a benchmark for short-horizon visual-motion dynamics in surgery, providing a unified evaluation platform for trustworthy surgical world models and laying a foundation for clinically safe intelligent surgical planning.

    \item We design a hierarchical protocol that shifts the focus from visual generation quality to trajectory-level dynamic consistency, assessing motion planning, closed-loop stability, and perturbation recovery to compensate for the lack of planning-oriented metrics in existing surgical world models.
    
    \item Our evaluation shows that existing surgical world models exhibit a clear mismatch between visual generation quality and trajectory accuracy, with limited motion precision and robustness, underscoring the necessity of planning-oriented assessment.

\end{itemize}

\section{Related Work}

\subsection{Surgical Motion Planning}

Surgical motion planning aims to model the motion of instruments within the local surgical field, providing an important foundation for robot-assisted surgical automation and intraoperative assistance. Existing studies can be broadly divided into two categories. One line of work recognizes surgical phases, actions, or workflow states from surgical videos, providing semantic context for high-level planning ~\cite{czempiel2020tecno, gao2021trans, nwoye2022rendezvous, yuan2022anticipation}. The other line leverages robotic kinematics, expert demonstrations, or historical trajectories to learn instrument end-effector motion, gesture transitions, or expert manipulation policies ~\cite{7805258, qin2020davincinet, shi2022recognition, hansen2026imitatecholec}. The former, represented by surgical phase recognition, action anticipation, and workflow prediction, has significantly improved models’ ability to understand surgical procedures and semantic states. The latter mainly focuses on gesture recognition, trajectory prediction, imitation learning, and skill assessment, providing an important basis for modeling robotic manipulation \cite{kim2024surgical, yuan2022anticipation}.

However, existing surgical motion planning ~\cite{shi2022recognition,hansen2026imitatecholec} and trajectory prediction methods ~\cite{yuan2021surgical, https://doi.org/10.1002/rcs.2441,liu2026sutureagent} often simplify the objective to predicting trajectory coordinates, robotic states, or action labels. Such methods can measure whether a model generates motion trajectories close to expert demonstrations, but they struggle to answer a more critical question for surgical planning: how the predicted trajectory affects the future surgical visual state. 
In other words, traditional trajectory prediction primarily models the motion trajectory, rather than jointly modeling instrument motion and future surgical video states. This limitation is particularly important for closed-loop surgical planning, where safe decision-making depends not only on whether a trajectory is smooth or close to expert behavior, but also on whether it leads to plausible instrument–tissue interactions, local state transitions, and changes in risk. Therefore, relying solely on trajectory prediction or kinematic modeling remains insufficient for planning evaluation in the context of surgical world modeling.

\subsection{Surgical World Models}

Recently, some studies have begun to explore surgical world models and surgical video generation. These methods typically use generative models or latent dynamics models to synthesize future surgical videos, and provide a degree of controllable generation through language prompts, reference frames, implicit actions, trajectory conditions, or other lightweight conditioning signals. Related studies suggest that world models can be used to generate future surgical scenes, augment rare action data, build interactive simulation environments, and even support robotic policy learning \cite{koju2025surgicalvision,rapuri2026saw,he2025surgworld, chen2025surgsora}. For example, Surgical Vision World Model \cite{koju2025surgicalvision} explores action-controllable data generation from unlabeled surgical videos; SAW \cite{rapuri2026saw} employs video diffusion models for trajectory-conditioned surgical action video synthesis; and SurgWorld \cite{he2025surgworld} attempts to infer pseudo-kinematics through generated videos and an inverse dynamics model to support surgical robot policy learning.

Although these works demonstrate the potential of surgical world models, existing evaluations mainly focus on the visual quality, temporal consistency, or downstream utility of generated videos. Common metrics such as FVD \cite{unterthiner2018towards} or CD-FVD \cite{ge2024content} measure whether generated videos are close to real videos in overall appearance and motion patterns by comparing their distributions in a pretrained video feature space \cite{unterthiner2018towards}. However, such metrics are difficult to use for determining whether the instrument trajectories in generated videos are accurate, and they cannot evaluate whether a model has truly learned local state transition dynamics that are useful for motion planning. A model may generate visually plausible videos while producing geometrically inaccurate instrument motion, trajectory drift during rollout, or dynamics inconsistent with the true consequences of surgical actions. Therefore, existing surgical world model studies still lack a benchmark specifically designed to evaluate trajectory accuracy, rollout stability, and perturbation robustness. SurgWMBench addresses this evaluation gap by examining whether surgical world models can not only generate visually plausible future scenes, but also accurately characterize instrument motion trajectories and maintain planning-oriented dynamic consistency under continuous prediction and perturbation conditions.

\section{Benchmark Design}

\begin{figure}[t]
    \centering
    \includegraphics[width=\linewidth]{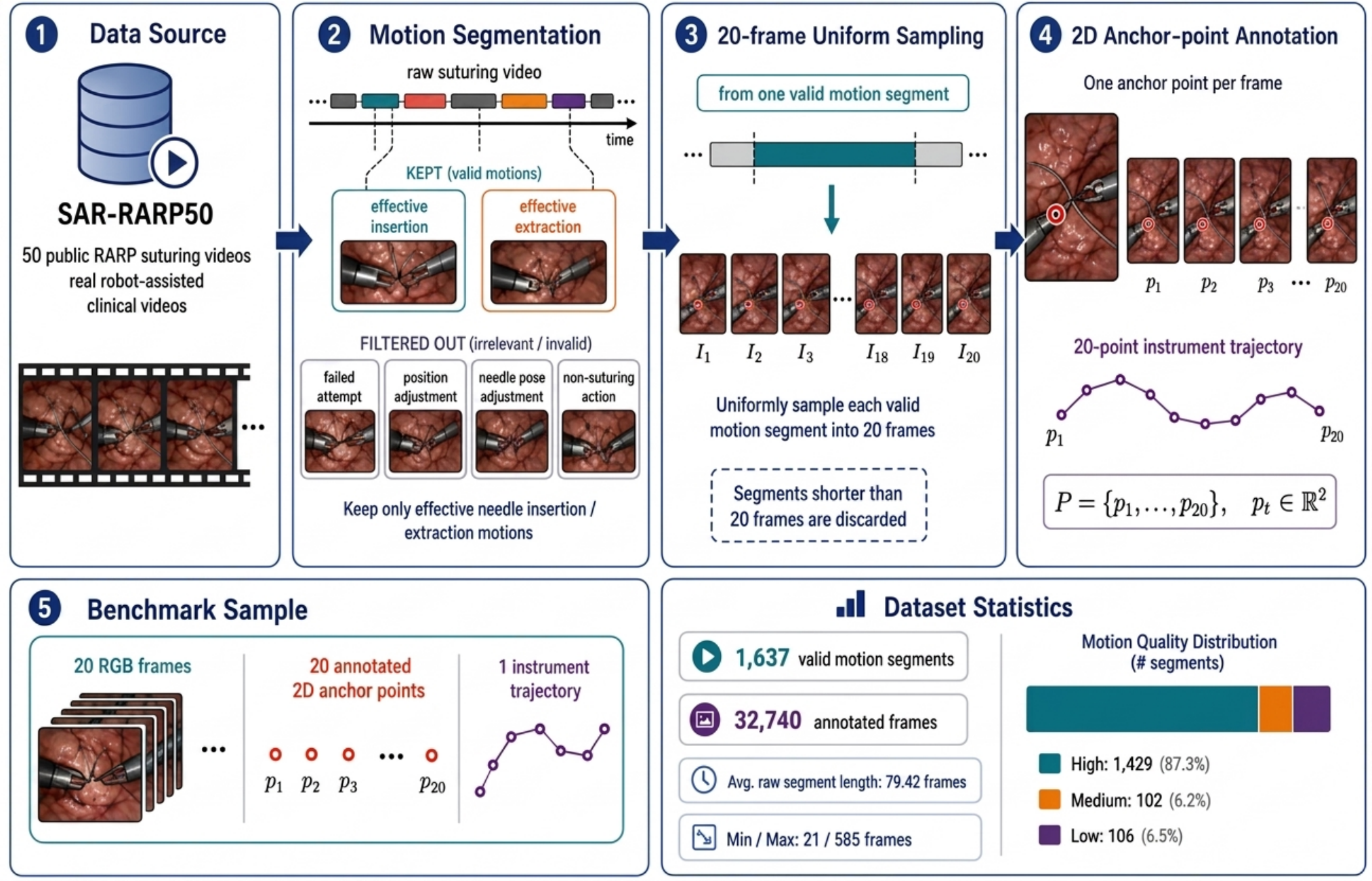}
    \caption{
Overview of the SurgWMBench construction pipeline. We segment valid needle insertion and extraction motions from SAR-RARP50 suturing videos, uniformly sample each motion segment into 20 RGB frames, and annotate one task-relevant 2D anchor point per frame. The resulting 20-point instrument trajectories serve as the core supervision signal for vision-based surgical motion planning.
    }
    \label{fig:benchmark_overview}
    \vspace{-2mm}
\end{figure}
\subsection{Overview}
SurgWMBench aims to provide a standardized evaluation platform for vision-based surgical motion planning, with a focus on short-horizon surgical dynamics. Unlike conventional surgical video benchmarks centered on phase recognition, action classification, or final task success rates, SurgWMBench focuses on whether a model can understand current surgical states from intraoperative images and predict the instrument motion trajectory over a short future horizon. This task formulation corresponds to a core problem in surgical world modeling: whether a model can learn the predictable relationship between the current visual state and future instrument motion.

Built upon the public SAR-RARP50 dataset~\cite{psychogyios2023sar}, which contains 50 real clinical robot-assisted RARP suturing videos with surgical action recognition and instrument segmentation annotations, SurgWMBench constructs a clinically grounded benchmark for vision-based surgical motion planning. The overall construction pipeline is shown in \autoref{fig:benchmark_overview}. From these videos, we define each effective needle insertion or extraction action as a motion segment (see Sec.\ref{subsec:motion_seg} for details), uniformly sample 20 frames from each valid segment, annotate one task-relevant 2D anchor point per frame, and connect the anchor points in temporal order to form a standardized 20-point instrument trajectory. This trajectory serves as the target for model prediction or planning. Through this design, SurgWMBench shifts the evaluation focus from discrete semantic prediction or generative quality to continuous motion modeling, enabling more direct assessment of local state transition modeling, continuous rollout stability, and recovery under perturbations.

After motion segmentation and removal of irrelevant clips, SurgWMBench contains 1,637 valid motion segments and 32,740 surgical images with 2D trajectory annotations. 
Clinicians further stratified the valid motion segments into three quality levels, resulting in 1,429 high-quality segments, 102 medium-quality segments, and 106 low-quality segments, corresponding to approximately 87.3\%, 6.2\%, and 6.5\% of the dataset, respectively. This distribution shows that most valid suturing motions have relatively clear visual trajectories, while the dataset also retains a set of complex and challenging samples for evaluating model performance under occlusion, locally complex trajectories, and visual uncertainty. More details about the dataset composition and statistics are provided in the Appendix~\ref{subsec:appendix_data_composition}.

\subsection{Motion Segmentation}
\label{subsec:motion_seg}

To adapt the original videos to the motion planning task, we first perform motion segmentation. A motion segment corresponds to an effective needle insertion or needle extraction action in a suturing motion. The starting point of a segment is defined as the moment when an effective insertion or extraction begins, and the ending point is defined as the moment when that insertion or extraction is completed. This definition restricts the annotation unit to a local operative process with clear dynamic meaning, allowing the model to focus on learning the relationship between instrument motion and local visual state changes.

During the segmentation process, we further remove clips that are irrelevant to the target action or unsuitable for dynamics modeling. The excluded cases mainly include unsuccessful suturing attempts, simple instrument position adjustments, needle pose adjustments, and other non-suturing-related actions. The purpose of this step is not to reduce task difficulty, but to minimize interference from invalid actions and individual manipulation strategies in the benchmark. By retaining effective needle insertion and extraction actions, SurgWMBench can more directly evaluate models’ ability to capture task-relevant short-horizon motion dynamics.

\subsection{Annotation Protocol and Interface}

Since the durations of different motion segments vary, we need to balance annotation cost and temporal coverage. Longer sequences can provide richer dynamic information but substantially increase manual annotation cost, whereas overly short sequences may not support effective motion modeling. Based on suggestions from clinical experts and annotation-cost considerations, we uniformly sample 20 frames from each effective motion segment and discard segments shorter than 20 frames. This strategy preserves the basic motion process while standardizing sample length for model training and evaluation.

For each sampled sequence, two annotators trained by clinical experts and experienced in robotic surgery video analysis use the customized annotation tool to perform two-dimensional trajectory annotation. In each frame, the annotators mark one task-relevant anchor point, and the 20 anchor points are connected in temporal order to obtain a 20-point polyline trajectory for the corresponding motion segment. This trajectory serves as the core ground-truth annotation of SurgWMBench.

For anchor point selection, we follow the principle of task relevance and trackability. The annotated point should, as much as possible, remain stably visible throughout the motion and reflect the overall direction of instrument movement. Its exact location may vary depending on the surgical scene and instrument visibility, but the goal is to obtain a continuous and trackable trajectory corresponding to the target suturing motion. This design avoids excessive dependence on a specific instrument structure or a single anatomical reference, while improving the applicability of the annotation protocol in complex clinical scenes.

\section{Experiments}

\subsection{Task Definition}
SurgWMBench aims to evaluate whether a model can predict short-horizon instrument trajectories based on intraoperative images and historical instrument motion. For each motion segment, we are given an image sequence $\mathcal{I} = \{ I_1, I_2, \ldots, I_{20} \}$ and the corresponding two-dimensional instrument trajectory $\mathcal{P} = \{{ p_1, p_2, \ldots, p_{20} }, \quad p_t \in \mathbb{R}^2\}$. In the main experiments, we use the first K frames as historical observations, and the model is required to predict the remaining $20-K$ trajectory points.

To systematically evaluate the model’s performance in terms of local state transition modeling, continuous rollout stability, and recovery under perturbations, we propose the following three tasks.

\textbf{Task 1: Short-horizon Prediction.} Short-horizon prediction task evaluates the model’s ability to capture local state transitions. Given the first K frames and their corresponding historical trajectory points, the model is required to predict the future 20 - K trajectory points:
\begin{equation}
    \hat{\mathcal{P}}_{K+1:20}
    =
    f
    \left(
        \mathcal{I}_{1:K},
        \mathcal{P}_{1:K}
    \right),
\end{equation}
where $\mathcal{I}_{1:K}$ denotes the real images from frame 1 to frame $K$, $\mathcal{P}_{1:K}$ denotes the first $K$ ground-truth trajectory points. This task corresponds to Local Transition Fidelity (LTF). It mainly examines whether the model can accurately predict the short-term motion trend of the instrument given the ground-truth historical state.

\textbf{Task 2: Autoregressive Rollout.} Autoregressive rollout task evaluates the stability of a model during continuous prediction. Unlike Task 1, this task does not only evaluate one-shot future trajectory prediction. Instead, starting from the first K frames and the first K ground-truth trajectory points, the model recursively feeds its own predictions as subsequent inputs to gradually generate the full future trajectory.

Specifically, we consider two rollout settings. The first setting is image-conditioned autoregressive rollout. In this setting, the model can access the real image sequence during rollout, but the future part of the trajectory history is recursively filled with the model’s own predictions. This setting is suitable for conventional trajectory prediction models and models that cannot generate future videos. It mainly evaluates whether trajectory prediction errors accumulate under recursive input. This setting can be formulated as:
\begin{equation}
\hat{p}_t = f \left( \mathcal{I}_{1:t},; \mathcal{P}_{1:K},; \hat{p}_{K+1}, \ldots, \hat{p}_{t-1} \right), \qquad t = K+1, \ldots, 20,
\end{equation}
where $\hat{p}_{\mathcal{\tau}}$ denotes the trajectory predicted by the model at time $\tau$.

The second setting is world-model rollout. In this setting, the model is required to generate both future visual states and future trajectories, and then use the generated video frames together with the predicted trajectories as subsequent rollout inputs. This setting is closer to the full surgical world modeling scenario and is suitable for models that jointly model future surgical video states and instrument trajectories. This setting can be formulated as:
\begin{equation}
\big( \hat{I}_t,\, \hat{p}_t \big) = f \left( \hat{I}_{t-1},\, \mathcal{P}_{1:K},\, \hat{p}_{K+1}, \ldots, \hat{p}_{t-1} \right), \quad t = K+1, \ldots, 20,
\end{equation}
where $\hat{I}_{\mathcal{\tau}}$ denotes the image generated by the world model. This task corresponds to Closed-loop Rollout Stability (CRS), which measures whether the model suffers from error accumulation and trajectory drift during continuous prediction.

\textbf{Task 3: Perturb-and-recover.} Perturb-and-recover evaluates the robustness of a model under input perturbations. This task is based on the short-horizon prediction setting in Task 1, but introduces perturbations into the historical trajectory input:
\begin{equation}
\hat{\mathcal{P}}_{K+1:20}
=
    f
    \left(
        \mathcal{I}_{1:K},
        \phi(\mathcal{P}_{1:K})
    \right),
\end{equation}
where $\phi$ denotes the perturbation operation. We consider Gaussian noise, which adds noise to historical trajectory points, and random masking, which removes or masks a subset of historical trajectory points. This task corresponds to Robustness under Perturbation and Shift (RPS), which examines whether the model can maintain stable predictions when historical motion information is incomplete or noisy.



\subsection{Evaluation Protocol and Metrics}
The evaluation protocol of SurgWMBench is organized around three complementary capabilities: local transition fidelity, closed-loop rollout stability, and robustness under perturbation and shift. Each capability corresponds to a specific task and a set of quantitative metrics.

\subsubsection{Local Transition Fidelity}
Local Transition Fidelity evaluates whether a model can accurately predict short-horizon local instrument motion. This capability is assessed through Task 1, namely short-horizon prediction under teacher forcing. We use Average Displacement Error (ADE) and Final Displacement Error (FDE) as the primary metrics.

(1) Average Displacement Error. ADE measures the average Euclidean distance between the predicted trajectory and the ground-truth trajectory over the entire prediction horizon:
\begin{equation}
\mathrm{ADE}
=
\frac{1}{20-K}
\sum_{t=K+1}^{20}
\left|
\hat{p}_t - p_t
\right|_2.
\end{equation}
(2) Final Displacement Error. FDE measures the Euclidean distance between the predicted endpoint and the ground-truth endpoint:
\begin{equation}
\mathrm{FDE}
=
\left|
\hat{p}_{20} - p_{20}
\right|_2.
\end{equation}
All coordinates are mapped back to the original image resolution before metric computation. Therefore, ADE and FDE are measured in pixels. ADE reflects the overall trajectory prediction accuracy, while FDE emphasizes the model’s ability to predict the final motion outcome.

\subsubsection{Closed-loop Rollout Stability}

Closed-loop Rollout Stability evaluates the stability of a model during continuous autoregressive prediction. This capability is assessed through Task 2, namely autoregressive rollout. Unlike short-horizon prediction, the model’s predictions are recursively used as subsequent inputs during rollout. Therefore, this task can reveal error accumulation that may not be observed under teacher forcing.

We use Rollout ADE@H as the primary metric. For a given horizon H, Rollout ADE@H is defined as the average prediction error from frame K+1 to frame K+H:
\begin{equation}
\mathrm{Rollout\ ADE@}H
=
\frac{1}{H}
\sum_{t=K+1}^{K+H}
\left|
\hat{p}_t - p_t
\right|_2.
\end{equation}
In the experiments, we report results under multiple horizons, such as $\mathrm{ADE@5}, \mathrm{ADE@10}, \mathrm{ADE@15}$. These metrics characterize how prediction error changes as the rollout horizon increases. If a model achieves low error at a short horizon but its error increases rapidly at longer horizons, this indicates that its local prediction ability cannot be stably transferred to continuous planning.

\subsubsection{Robustness under Perturbation and Shift}
\label{subsec:robustness}
Robustness under Perturbation and Shift evaluates whether a model can maintain stable predictions when the input trajectory is perturbed. This capability is assessed through Task 3, namely perturb-and-recover. We test the model under Gaussian noise and random masking conditions, and compare the results with those under clean input. The primary metric is Performance Drop Rate (PDR). Let $E_{\mathrm{clean}}$ denote the ADE under clean input, and $E_{\mathrm{perturbed}}$ denote the ADE under perturbed input. PDR is defined as:
\begin{equation}
\mathrm{PDR}
=
\frac{
E_{\mathrm{perturbed}} - E_{\mathrm{clean}}
}{
E_{\mathrm{clean}}
}
\times 100\%.
\end{equation}
A lower PDR indicates stronger robustness to input perturbations. This metric reflects whether a model over-relies on precise historical trajectory points, as well as its ability to recover when historical motion information is incomplete or noisy.

\subsection{Experimental Results}

We compare four world model baselines under two training settings. The first setting evaluates \emph{pure video generation}, where the model predicts future visual states without explicit trajectory supervision. The second setting evaluates \emph{joint image generation and trajectory prediction}, where the same backbone is augmented with a trajectory prediction head and optimized with both visual and motion supervision. This comparison is designed to reveal whether explicit motion supervision improves trajectory prediction while preserving the fidelity of generated surgical frames.

Table~\ref{tab:main_results} merges the two settings into a single table for direct row-wise comparison. Columns under group (a) report the pure image-generation setting with SSIM, PSNR, and LPIPS. Columns under group (b) report the joint image-generation and trajectory-prediction setting with the same three image-quality metrics together with ADE and FDE. This layout keeps the comparison compact while still making the two experimental settings easy to distinguish.

\begin{table*}[t]
    \centering
    \footnotesize
    \caption{Main experimental results on SurgWMBench. Group (a) reports pure image generation, and group (b) reports joint image generation with trajectory prediction. Higher is better for SSIM and PSNR, while lower is better for LPIPS, ADE, and FDE.}
    \label{tab:main_results}
    {\renewcommand{\arraystretch}{1.12}
    \setlength{\tabcolsep}{4.5pt}
    \begin{tabular}{lcccccccc}
        \toprule
        & \multicolumn{3}{c}{\textbf{(a) Video Generation}} & \multicolumn{5}{c}{\textbf{(b) Joint Video Generation + Motion Planning}} \\
        \cmidrule(lr){2-4} \cmidrule(lr){5-9}
        Baseline & PSNR $\uparrow$ & SSIM $\uparrow$ & LPIPS $\downarrow$ & PSNR $\uparrow$ & SSIM $\uparrow$ & LPIPS $\downarrow$ & ADE $\downarrow$ & FDE $\downarrow$ \\
        \midrule
        VideoGPT~\cite{yan2021videogpt} & 13.85 & \textbf{0.58} & 0.68 & 13.01 & 0.53 & 0.69 & 52.61 & 70.57 \\
        iVideoGPT~\cite{wu2024ivideogpt} & 13.87 & 0.53 & 0.63 & 13.59 & 0.52 & 0.62 & \textbf{51.32} & \textbf{65.11} \\
        HieraSurg~\cite{biagini2025hierasurg} & \textbf{16.88} & \textbf{0.58} & \textbf{0.52} & \textbf{16.74} & \textbf{0.56} & \textbf{0.52} &  176.20 & 174.70  \\
        SurgSora~\cite{chen2025surgsora} & 6.20 & 0.33 & 0.64 & 7.26 & 0.36 & 0.65 & 64.12 & 81.54\\
        \bottomrule
    \end{tabular}}
\end{table*}

To further evaluate robustness, we perturb the historical trajectory input and re-evaluate the jointly trained models. Following the protocol in Sec.\ref{subsec:robustness}, we consider two perturbation types: Gaussian noise and random masking. We report the corresponding Performance Drop Rate (PDR), computed from the ADE change between clean and perturbed inputs. A lower PDR indicates that a model is less sensitive to imperfect historical motion signals and is therefore more reliable for practical surgical motion planning scenarios.

\begin{table*}[t]
    \centering
    \footnotesize
    \caption{Robustness under trajectory perturbation. We report clean-input ADE and FDE once for each baseline, and then report perturbed ADE and FDE under Gaussian noise and random masking. Lower is better for all metrics.}
    \label{tab:perturbation_results}
    {\renewcommand{\arraystretch}{1.10}
    \setlength{\tabcolsep}{4.2pt}
    \begin{tabular}{lcccccc}
        \toprule
        & \multicolumn{2}{c}{\textbf{Clean}} & \multicolumn{2}{c}{\textbf{Gaussian}} & \multicolumn{2}{c}{\textbf{Mask}} \\
        \cmidrule(lr){2-3} \cmidrule(lr){4-5} \cmidrule(lr){6-7}
        Baseline & ADE $\downarrow$ & FDE $\downarrow$ & ADE $\downarrow$ & FDE $\downarrow$ & ADE $\downarrow$ & FDE $\downarrow$ \\
        \midrule
        VideoGPT~\cite{yan2021videogpt} & 52.61 & 70.57 & 141.20 & 157.12 & 131.14 & 146.26  \\
        iVideoGPT~\cite{wu2024ivideogpt} & \textbf{51.32} & \textbf{65.11} & 131.34 & 152.01 & \textbf{107.34} & \textbf{113.75}  \\
        HieraSurg~\cite{biagini2025hierasurg} & 176.20 & 174.70 & 176.14 & 174.85 & 176.23 & 174.81  \\
        SurgSora~\cite{chen2025surgsora} & 64.12 & 81.54 & \textbf{104.33} & \textbf{115.73} & 123.21 & 140.78 \\
        \bottomrule
    \end{tabular}}
\end{table*}

\begin{figure}
    \centering
    \includegraphics[width=0.98\linewidth]{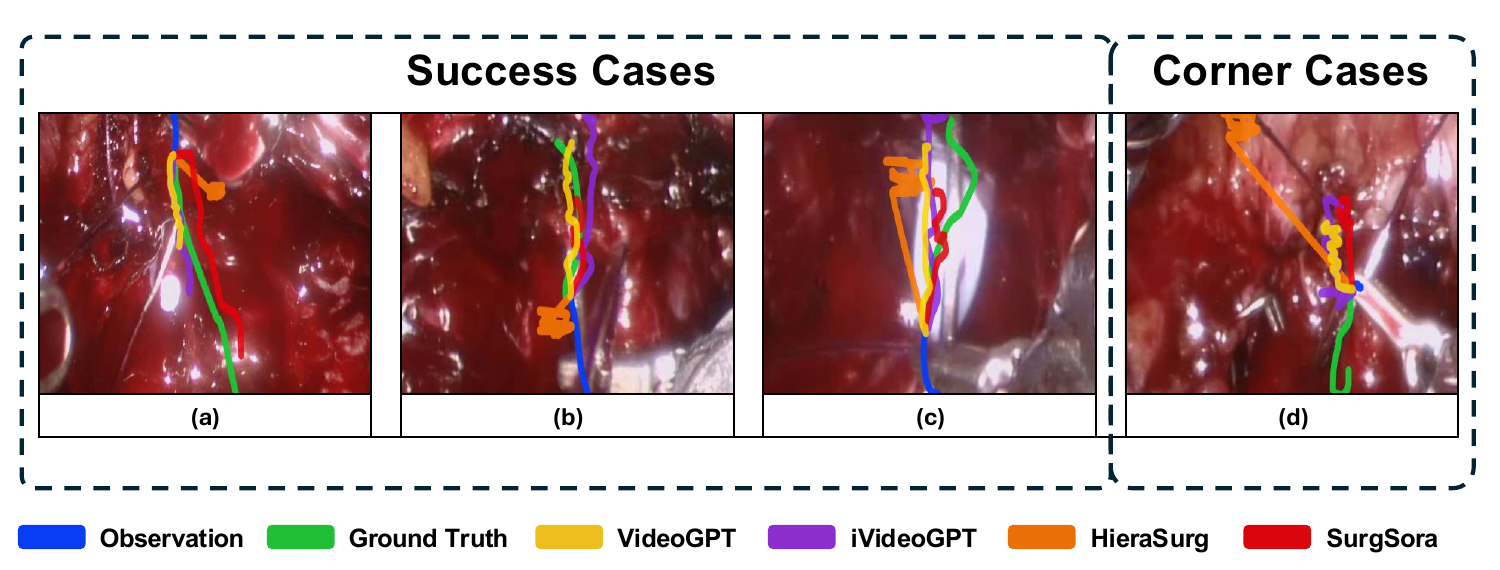}
    \caption{Qualitative trajectory prediction results on SurgWMBench. The blue curves denote observed historical instrument trajectories, and the green curves denote ground-truth future trajectories. The brown, purple, orange, and red curves denote predictions from VideoGPT, iVideoGPT, HieraSurg, and SurgSora, respectively. Panels (a)--(c) show representative success cases where several baselines follow the overall future motion direction, while panel (d) shows a corner case with substantial trajectory deviation under complex local visual conditions.}
    \label{fig:qualitative_results}
\end{figure}

\subsection{Results Analysis}

The results show that visual generation quality does not ensure planning-oriented motion accuracy. As shown in Table~\ref{tab:main_results}, HieraSurg achieves the strongest image-level performance in the joint setting, with the highest PSNR and lowest LPIPS, but its trajectory prediction error remains substantially larger than those of VideoGPT and iVideoGPT. In contrast, iVideoGPT obtains the best motion prediction performance with the lowest ADE and FDE, although its visual generation metrics are not the best among all baselines. This discrepancy supports SurgWMBench's motivation. Generation-oriented metrics alone are insufficient for assessing whether a surgical world model has learned instrument motion dynamics for planning. The qualitative results in Figure~\ref{fig:qualitative_results} provide consistent evidence. In the success cases, some predicted trajectories can follow the coarse direction of the ground-truth future motion, but different models still exhibit visible differences in geometric alignment, endpoint accuracy, and local trajectory shape. In the corner case, predictions deviate substantially from the ground truth, suggesting that complex local visual context, occlusion, or ambiguous instrument-tissue interaction can still challenge current world-model baselines.

The perturbation results in Table~\ref{tab:perturbation_results} show that baselines remain sensitive to imperfect motion histories. VideoGPT and iVideoGPT achieve strong clean-input trajectory prediction, but their errors increase sharply under Gaussian noise and random masking, suggesting limited recovery from noisy or incomplete histories. HieraSurg shows little performance change under perturbations, but this should be interpreted alongside its high clean-input error, since input insensitivity does not imply reliable planning. Overall, quantitative and qualitative results indicate that current surgical world models remain limited in trajectory accuracy, robustness, and planning-oriented dynamics. Additional rollout analysis is provided in Appendix~\ref{subsec:rollout_analysis}, where we further examine error accumulation during continuous autoregressive prediction.
\section{Conclusion}

We introduced SurgWMBench, a vision-based benchmark for evaluating short-horizon surgical world modeling and instrument motion planning. Built from real robot-assisted suturing videos with annotated 2D trajectories, SurgWMBench assesses trajectory accuracy, closed-loop rollout stability, and robustness to perturbed motion inputs, revealing that visual generation quality does not necessarily imply accurate motion prediction. Future work will broaden SurgWMBench to more procedures, instruments, and institutions, incorporate tissue-interaction representations and extend evaluation toward longer-horizon action-conditioned planning under clinical and robotic safety constraints.

\bibliography{ref}
\bibliographystyle{plain}

\clearpage

\appendix

\section{Appendix}

\subsection{Principles for selecting data sources and designing benchmark}

We choose SAR-RARP50 as the data source for three main reasons. First, it is collected from real robot-assisted surgical scenarios. Compared with simulator, bench-top, or ex-vivo experimental data, it better reflects the visual complexity, instrument interactions, and surgeon-specific variations in clinical environments. Second, the dataset focuses on the suturing phase in RARP, containing a large number of instrument motions with clear objectives and continuous motion structures. Third, suturing is highly dependent on the local dynamic relationship among the needle, instruments, and tissue, making it suitable for evaluating whether models can learn short-horizon surgical dynamics.

For the benchmark, the design of SurgWMBench follows the following three principles:
\begin{itemize}
    \item First, the data should come from real-world robot-assisted surgical scenarios to preserve as much of the visual complexity and motion variability as possible in clinical procedures.
    \item Second, the task should focus on short-term effective motions within a specific task process to reduce interference from irrelevant operations and individual strategy differences on dynamic modeling.
    \item Third, annotations should be presented in the form of reproducible and quantifiable trajectories to enable fair comparisons between different models under a unified protocol.

\end{itemize}

\subsection{Experimental Setup}
\textbf{Dataset Split.} We construct training, validation, and test sets based on SurgWMBench. To avoid similar segments from the same surgical video appearing in both the training and test sets, all splits are performed at the video level rather than randomly at the segment level. Therefore, the training videos and test videos are completely separated, preventing video-level contextual leakage. The training, validation, and test splits will follow a unified protocol in the final released version and will be reported with the ratio of 7/1.5/1.5.

\textbf{Input Preprocessing.} All models use RGB images and two-dimensional instrument trajectories as inputs. We do not use the original segmentation masks, action labels, or other auxiliary annotations from SAR-RARP50. 

Trajectory coordinates are not normalized during training or evaluation. Since image resizing may change the coordinate scale, all predicted trajectories and ground-truth trajectories are mapped back to the original image resolution before computing ADE, FDE, Rollout ADE, and PDR. This ensures that all geometric errors are computed in the original pixel space and have a consistent physical interpretation.

\textbf{Training Protocol.} To ensure fair comparison, all baselines use the same training, validation, and test splits. Models can only access RGB images and two-dimensional trajectory annotations. Segmentation masks, action labels, and quality labels are not allowed as training supervision. Quality labels are used only for subsequent difficulty-stratified performance analysis.


\textbf{Compute Resources.} All experiments were conducted on a workstation equipped with three NVIDIA RTX PRO 6000 Blackwell Workstation Edition GPUs, each with 96 GiB of GPU memory. The software environment was based on CUDA 13 and PyTorch 2.11.

\subsection{Customized annotation interface for SurgWMBench}
\begin{figure}[t]
    \centering
    \includegraphics[width=\linewidth]{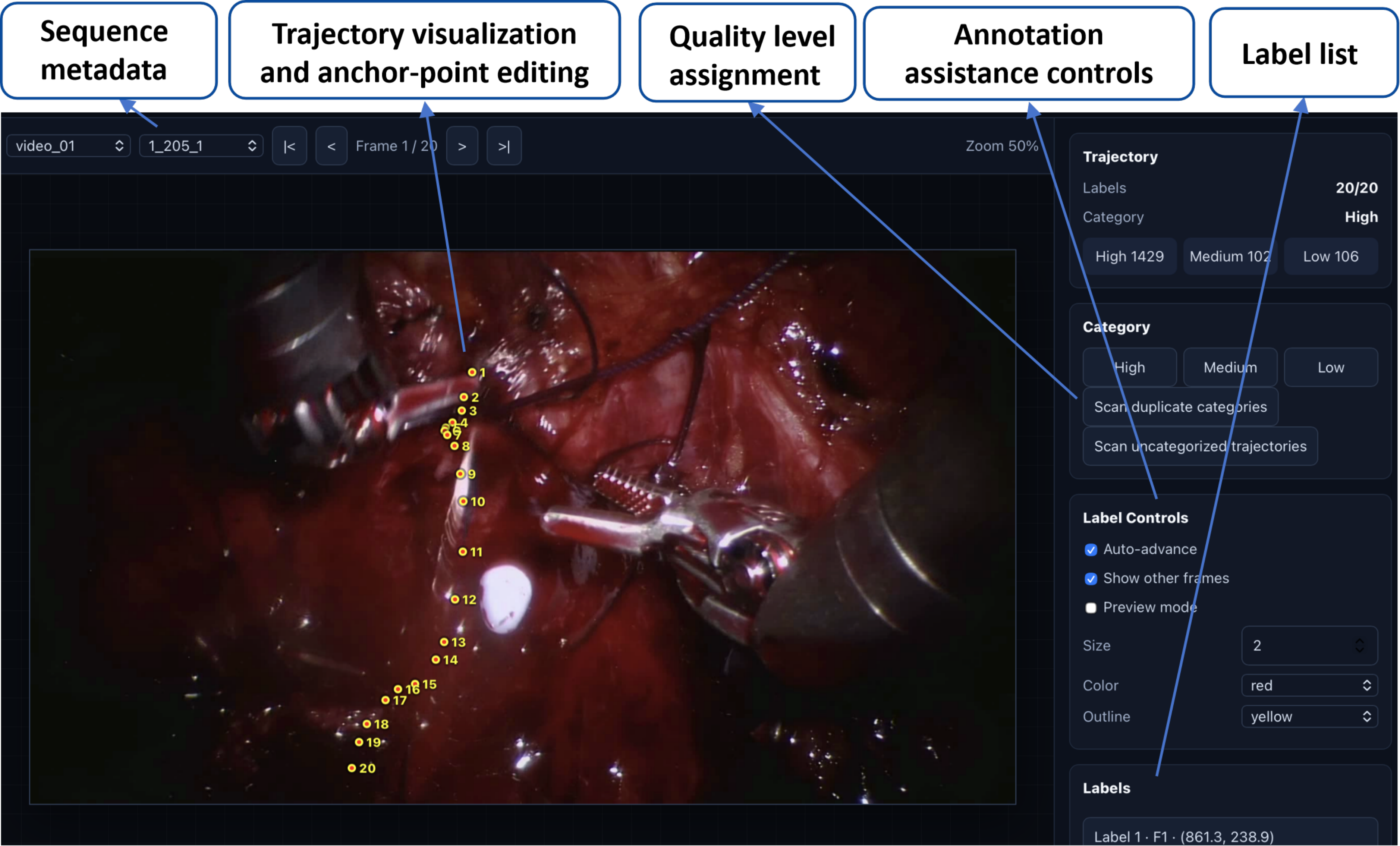}
    \caption{Customized annotation interface for SurgWMBench. }

    \label{fig:interface}
    \vspace{-2mm}
\end{figure}

To improve annotation efficiency and consistency, we develop a customized annotation interface, as shown \autoref{fig:interface}. Annotators can view the target image sequence and its basic information, including the case ID and trajectory sequence ID. Annotators can interactively select, adjust, delete anchor points, and assign quality levels. This interface ensures that all samples follow a unified annotation format and reduces the operational cost of long-sequence trajectory annotation.

\subsection{Interpolation Strategy}
\label{subsec:interpolation_strategy}

Each clip in SurgWMBench contains dense real video frames but only 20 human-annotated coordinate anchors. To increase the number of effective training samples, we first densify the sparse trajectory annotations and then construct overlapping training clips with a sliding-window strategy. Given the human anchors
\[
\mathcal{A}=\{(t_i, \mathbf{p}_i)\}_{i=1}^{20}, \quad \mathbf{p}_i=(x_i,y_i),
\]
where $t_i$ is the local frame index and $\mathbf{p}_i$ is the annotated 2D pixel coordinate, we estimate pseudo coordinates $\hat{\mathbf{p}}_t$ for non-anchor frames by interpolating the $x$ and $y$ coordinates independently along the temporal axis.

The dataset provides four interpolation methods: \texttt{linear}, \texttt{pchip}, \texttt{akima}, and \texttt{cubic\_spline}. Specifically, \texttt{linear} performs piecewise linear interpolation; \texttt{pchip} uses shape-preserving piecewise cubic Hermite interpolation; \texttt{akima} applies Akima interpolation to reduce oscillations caused by abrupt local changes; and \texttt{cubic\_spline} fits a smooth cubic spline trajectory. We use \texttt{linear} interpolation as the default setting due to its conservative behavior between adjacent human anchors, while the other methods are retained as alternative dense pseudo-label generation strategies. Anchor frames always keep their original human annotations, whereas non-anchor frames are treated as interpolated pseudo labels rather than human ground truth.

After dense pseudo coordinates are obtained, we apply a sliding window over each clip. For a window length $L$ and stride $s$, the $k$-th training sample is defined as
\[
\mathcal{W}_k=\{(I_t,\hat{\mathbf{p}}_t)\}_{t=ks}^{ks+L-1},
\]
where $I_t$ is the real surgical video frame and $\hat{\mathbf{p}}_t$ is either a human annotation or an interpolated pseudo coordinate. Thus, a clip with $T$ frames produces
\[
N=\left\lfloor \frac{T-L}{s} \right\rfloor + 1
\]
training samples when $T\geq L$. This strategy substantially increases the number of temporal training samples while preserving the original sparse human annotations as the primary supervision signal.

\subsection{Dataset Composition and Statistics}
\label{subsec:appendix_data_composition}

SurgWMBench is constructed from the 50 public videos in SAR-RARP50. After motion segmentation and removal of irrelevant clips, we obtain 1,637 valid motion segments. Each segment contains 20 frames and corresponds to an instrument trajectory composed of 20 two-dimensional anchor points. In total, SurgWMBench contains 32,740 surgical images with two-dimensional trajectory annotations.

In terms of segment duration, the average length of the original valid segments before sampling is 79.42 frames, with the shortest segment containing 21 frames and the longest segment containing 585 frames. This indicates that the dataset includes both relatively short and direct needle insertion or extraction motions, as well as longer suturing segments with more complex motion processes. All segments are finally uniformly sampled into 20 frames to form a consistent benchmark input format.

\begin{figure}[t]
    \centering
    \includegraphics[width=0.9\linewidth]{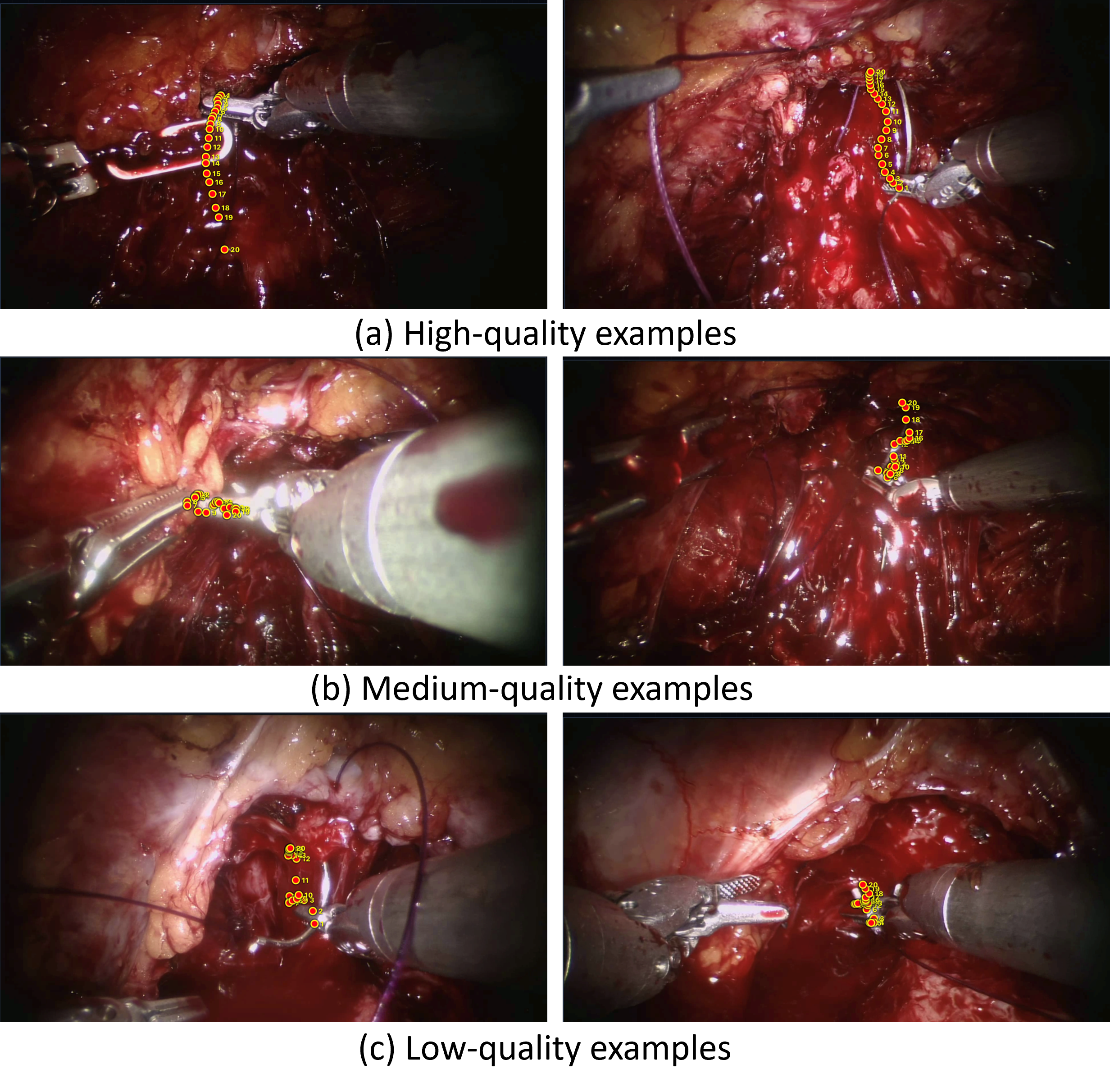}
    \caption{Examples of SurgWMBench annotation with different quality levels.
 }
    \label{fig:example}
    \vspace{-2mm}
\end{figure}

According to the quality levels, SurgWMBench contains 1,429 high-quality segments, 102 medium-quality segments, and 106 low-quality segments. The overall proportions are approximately 87.3\%, 6.2\%, and 6.5\%, respectively. This distribution shows that most valid suturing motions have relatively clear visual trajectories, while the dataset also retains a certain number of complex and challenging samples to evaluate model performance under occlusion, locally complex trajectories, or visual uncertainty. \autoref{fig:example} shows representative examples from different quality levels, where the annotated anchor points remain clearly trackable in high-quality samples but become progressively more challenging under occlusion, complex interactions, or uncertain local motion.

It should be emphasized that SurgWMBench does not use the original segmentation masks, action labels, or other auxiliary annotations from SAR-RARP50 as model inputs or supervision targets. Our goal is to construct a benchmark focused on image-based motion planning; therefore, the core supervision signal comes only from the newly annotated two-dimensional instrument trajectories. The training, validation, and test splits will follow a unified protocol in the final released version and will be reported with the ratio of 7/1.5/1.5.

\subsection{Additional Analysis of Closed-loop Rollout Stability}
\label{subsec:rollout_analysis}

Table~\ref{tab:rollout_results} reports the additional closed-loop rollout results under different prediction horizons. Across all baselines, both ADE and FDE increase as the rollout horizon extends from @5 to @15, indicating that prediction errors accumulate during autoregressive inference. This confirms that local short-horizon prediction accuracy under teacher forcing does not directly guarantee stable continuous planning. Even models with competitive clean-input performance still suffer from progressively larger deviations when their own predictions are recursively used as future inputs.

Among the evaluated baselines, iVideoGPT achieves the lowest rollout errors across all reported horizons, followed closely by VideoGPT. This suggests that these two models better preserve short-term motion consistency during continuous prediction. SurgSora shows moderate rollout performance but still exhibits clear error growth at longer horizons, especially in FDE, reflecting endpoint drift. HieraSurg suffers from large rollout errors, which is consistent with its weaker trajectory prediction performance in Table~\ref{tab:main_results}. Overall, the rollout results reveal that current surgical world model baselines remain limited in closed-loop stability, highlighting the need for motion-aware training objectives and evaluation protocols that explicitly penalize autoregressive error accumulation.

\begin{table*}[t]
    \centering
    \footnotesize
    \caption{Additional closed-loop rollout results on SurgWMBench. ADE and FDE are reported at rollout horizons of 5, 10, and 15 future steps, with all errors measured in the original image pixel space. Lower values indicate better autoregressive trajectory accuracy and stronger rollout stability.}
    \label{tab:rollout_results}
    {\renewcommand{\arraystretch}{1.12}
    \setlength{\tabcolsep}{4.5pt}
    \begin{tabular}{lcccccc}
        \toprule
        & \multicolumn{3}{c}{\textbf{ADE $\downarrow$}} & \multicolumn{3}{c}{\textbf{FDE $\downarrow$}} \\
        \cmidrule(lr){2-4} \cmidrule(lr){5-7}
        Baseline & @5 & @10 & @15 & @5 & @10 & @15  \\
        \midrule
        VideoGPT~\cite{yan2021videogpt} & 52.6 & 76.1 & 99.4 & 70.5 & 118.9 & 162.4 \\
        iVideoGPT~\cite{wu2024ivideogpt} & \textbf{51.3} & \textbf{71.8} & \textbf{94.6} & \textbf{65.1} & \textbf{111.8} & \textbf{156.3} \\
        HieraSurg~\cite{biagini2025hierasurg} & 64.1 & 84.1 & 108.6 & 81.5 & 121.3 & 176.2 \\
        SurgSora~\cite{chen2025surgsora} & 176.2 & 178.3 & 188.0 & 174.7 & 185.3 & 223.1\\
        \bottomrule
    \end{tabular}}
\end{table*}



\end{document}